\documentclass[runningheads]{llncs}
\usepackage[T1]{fontenc}

\usepackage{graphicx,verbatim}
\usepackage{orcidlink}

\usepackage{amsmath,amssymb}
\usepackage{adjustbox}
\usepackage{hyperref}

\usepackage[nolist]{acronym}
\usepackage{ragged2e}
\usepackage{tabularx}
\usepackage{ulem}
\usepackage{graphicx}
\usepackage{array,multirow,graphicx}
\usepackage{xspace}
\usepackage[capitalise]{cleveref}

\usepackage{pgfplots}
\pgfplotsset{compat=1.18}
\usepackage{pgfplotstable}

\makeatletter
\newcommand*{\org@overidelabel}{}
\let\org@overridelabel\@verridelabel
\@ifpackagelater{acronym}{2015/03/21}{
  \renewcommand*{\@verridelabel}[1]{%
    \@bsphack
    \protected@write\@auxout{}{\string\AC@undonewlabel{#1@cref}}%
    \org@overridelabel{#1}%
    \@esphack
  }%
}{
  \renewcommand*{\@verridelabel}[1]{%
    \@bsphack
    \protected@write\@auxout{}{\string\undonewlabel{#1@cref}}%
    \org@overridelabel{#1}%
    \@esphack
  }%
}
\makeatother

\newcommand{\mytitle}{Faster, Self-Supervised Super-Resolution for Anisotropic Multi-View MRI Using a Sparse Coordinate Loss}

\begin{document}
\def\bmvaOneDot{.}

\def\eg{{e.g}\bmvaOneDot}
\def\ie{{i.e}\bmvaOneDot}
\def\Ie{{i.e}\bmvaOneDot}
\def\Eg{{E.g}\bmvaOneDot}
\def\etal{{et al}\bmvaOneDot}

\title{\mytitle}
\titlerunning{Self-Supervised Multi-View Super-Resolution}

\author{Maja Schlereth\inst{1}\orcidlink{0009-0003-4418-7065} \and
Moritz Schillinger\inst{1}\orcidlink{0009-0008-3506-6719} \and
Katharina Breininger\inst{1,2}\orcidlink{0000-0001-7600-5869}}

\authorrunning{M. Schlereth et al.}

\institute{Department Artificial Intelligence in Biomedical Engineering, Friedrich-Alexander-Universität Erlangen-Nürnberg, Erlangen, Germany \\
\email{maja.schlereth@fau.de}
\and
Center for AI and Data Science (CAIDAS), Julius-Maximilians-Universität Würzburg, Würzburg, Germany
}

\maketitle              
\begin{abstract}
Acquiring images in high resolution is often a challenging task. Especially in the medical sector, image quality has to be balanced with acquisition time and patient comfort. To strike a compromise between scan time and quality for \ac{mr} imaging, two anisotropic scans with different \ac{lr} orientations can be acquired. 
Typically, LR scans are analyzed individually by radiologists, which is time consuming and can lead to inaccurate interpretation. 
To tackle this, we propose a novel approach for fusing two orthogonal anisotropic \ac{lr} \ac{mr} images to reconstruct anatomical details in a unified representation. Our multi-view neural network is trained in a self-supervised manner, without requiring corresponding \ac{hr} data. To optimize the model, we introduce a sparse coordinate-based loss, enabling the integration of \ac{lr} images with arbitrary scaling.
We evaluate our method on \ac{mr} images from two independent cohorts. Our results demonstrate comparable or even improved \ac{sr} performance compared to \ac{sota} self-supervised \ac{sr} methods for different upsampling scales. By combining a patient-agnostic offline and a patient-specific online phase, we achieve a substantial speed-up of up to ten times for patient-specific reconstruction while achieving similar or better \ac{sr} quality. Code is available at \url{https://github.com/MajaSchle/tripleSR}.

\end{abstract}

\section{Introduction}
\acresetall
\Ac{mr} imaging allows the 3-D assessment of bone and soft tissue anomalies without ionizing radiation. To increase patient comfort and minimize motion artifacts, it is important to keep image acquisition times as low as reasonably possible~\cite{Riek.1995,Zaitsev.2015}. One important factor that impacts acquisition time is the inter-slice (out-of-plane) resolution.
Acquiring multiple anisotropic 3-D images with a high in-plane resolution can effectively reduce and split up scanning time~\cite{Thrower.2021}. Conventionally, the individually acquired images then need to be assessed separately, which is time consuming and error prone as lesions might only be visible in one of the scans, potentially leading to misinterpretations. 
Multi-view \ac{sr} enables the generation of a single \ac{hr} image given several \ac{lr} images for a simpler diagnostic process \cite{Jia.2017}.

In this work, we introduce a novel self-supervised \ac{mr} \ac{sr} method that combines pretraining on a small, anisotropic dataset and per-patient adaptation during inference. Incorporating multi-patient information enables the use of cross-patient similarities while maintaining applicability to clinical scenarios with limited data availability. Accordingly, we split our optimization process into two phases: The ``offline phase'' which constitutes patient-agnostic feature extraction, and the ``online phase'' for additional patient-specific adaptation and subsequent inference to generate the final \ac{sr} image. Minimizing runtimes for the online phase is critical for efficient deployment in production settings, such as hospitals, where fast image processing is essential. Specifically, we incorporate two orthogonal anisotropic \ac{lr} \ac{mr} scans to form a joint representation that facilitates the generation of accurate \ac{sr} images. 
Following clinical routine, where typically no \ac{hr} data is available, we train our network only on the available \ac{lr} data using a fully self-supervised approach, circumventing the need for \ac{hr} images as reference for training. 
By integrating a sparse coordinate-based loss function, we are able to 1) use images with varying resolution for training and 2) allow for arbitrary-scale upsampling.
For evaluation, we perform extensive evaluations on two publicly available datasets~\cite{Bakas.2017,Menze.2015,vanEssen.2013} and demonstrate that our proposed method generalizes across datasets and to unseen \ac{mr} sequences. We specifically highlight that our approach speeds up the \ac{sr} process substantially compared to \ac{sota} self-supervised \ac{sr} methods while preserving image quality.

\subsection{Related Work}
Several methods have been proposed to improve the resolution of \ac{mr} image data~\cite{Zhao.2021,Chen.2018,Nasrollahi.2014,Ooi.2021}. They can be divided into \ac{sisr} and \ac{misr} methods~\cite{Zhao.2019,Ooi.2021}. 
\Ac{misr} utilizes different (anisotropic) views or several sequentially acquired \ac{lr} scans by merging them to generate a higher-resolution image. 

For \ac{mr} \ac{sr} specifically, multiple learning-based methods have been proposed that utilize matching \ac{lr} and \ac{hr} data as reference for training \ac{sr} models.
Lyu \etal~\cite{Lyu.2023} performed \ac{mr} \ac{sr} by matching features of the target \ac{lr} contrast and the available reference \ac{hr} image of a different domain. They used a transformer with a texture-preserving branch and contrastive learning to enhance the textural details of the \ac{sr} image.
To overcome the issue of generalization of \ac{sr} approaches to different scales, Tan \etal~\cite{Tan.2020} proposed an arbitrary scale \ac{sr} approach for brain \ac{mr} images. They combined a Weight Prediction Network with SRGAN and achieved comparable performance to \ac{sota} \ac{sr} techniques. However, their approach relies on a large training dataset with \ac{hr} images and they performed evaluation only on 2-D slices. 
Recently, \acp{inr} have become more prominent in the context of image \ac{sr}. In the work of Wu \etal~\cite{Wu.2023}, an \ac{inr} for \ac{mr} image \ac{sr} was proposed. This method allows implicit modeling of a continuous function of sparsely sampled data points and projects the learned intensity distribution onto a more densely sampled grid. They achieved high reconstruction performance for several upsampling scales but still relied on paired isotropic \ac{lr} and \ac{hr} data. 
Distinct from the previously presented methods, other approaches allow for reference-free training of \ac{sr} network while integrating arbitrary-scale upsampling. McGinnis \etal~\cite{McGinnis.2023} used two anisotropic \ac{lr} \ac{mr} images of different \ac{mr} sequences to perform \ac{inr}-based \ac{sr}. While achieving very good \ac{sr} reconstruction, their approach requires patient-specific training in the order of 10 to 20 minutes per image, which may not be clinically feasible. 

\section{Methodology}

\begin{figure}[!ht]
\includegraphics[width=\textwidth]{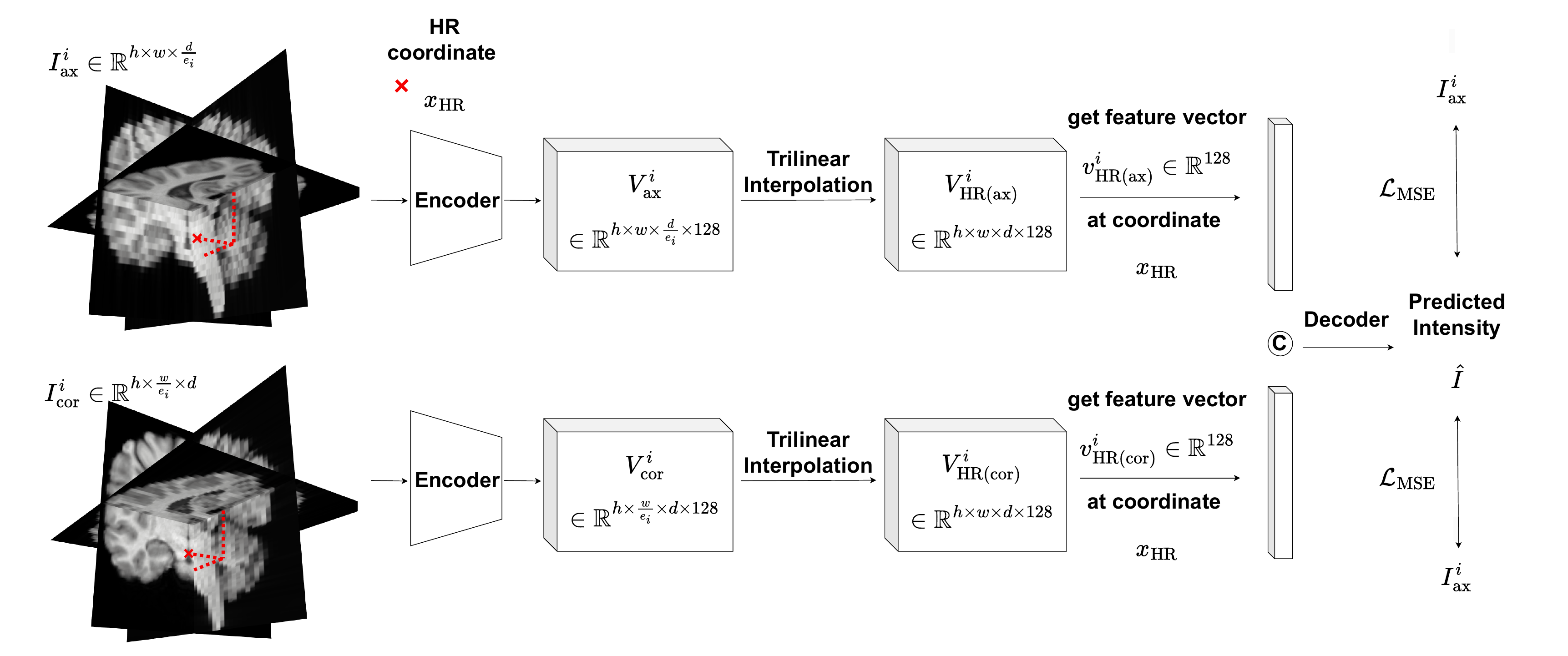}
\caption{Overview of the network architecture proposed in this study.
} \label{network}
\end{figure}

Our model takes $N$ paired \ac{lr} images $I_\text{ax}^i$ and $I_\text{cor}^i$ from a training set $T = \{ I_\text{ax}^i \in \mathbb{R}^{h\times w\times \frac{d}{e_i}}, I_\text{cor}^i \in \mathbb{R}^{h\times \frac{w}{e_i}\times d} \rbrace_{i=1}^{N}$ as inputs, with $e_i$ being the upsampling scale of the $i$-th \ac{lr}-image pair. In \cref{network} the complete training process is shown. A convolutional encoder extracts feature maps $V_\text{ax}^i  \in \mathbb{R}^{h\times w\times \frac{d}{e_i}\times 128} $ and $V_\text{cor}^i \in \mathbb{R}^{h\times \frac{w}{e_i}\times d \times 128}$ which are subsequently used for reconstruction. We adapt a \ac{rdn}~\cite{Zhang.2018}, which is specifically developed for image \ac{sr}, by removing the upsampling layer and applying 3-D convolution. 

Each element in the feature map corresponds to a voxel in the respective \ac{lr} image. In the \ac{hr} image, coordinates may lie between those of the \ac{lr} counterparts, along with the corresponding voxel data. To obtain a feature vector at a corresponding high-resolution coordinate $x_\text{HR}$, we perform trilinear interpolation on the feature maps $V_\text{ax}^i$ and $V_\text{cor}^i$ and generate two \ac{hr} feature maps with the shape $h\times w\times d\times 128$ each. 
From this, we sample the feature vectors $v_\text{HR(ax)}^i$ and $v_\text{HR(cor)}^i$ at a specific \ac{hr} coordinate $x_\text{HR}$. The two extracted feature vectors $v_\text{HR(ax)}^i$ and $v_\text{HR(cor)}^i$ corresponding to the \ac{hr} coordinate are concatenated and fed to the decoder. The decoder consists of eight fully connected layers each followed by ReLU activation with a residual connection after the fourth layer. Each inner layer has 512 input and output features. The decoder outputs the predicted voxel intensity $\hat{I}$ of the respective spatial coordinate $x_\text{HR}$, following~\cite{Wu.2023}.

We split our \ac{sr} process into two phases. The first part is a training phase on a disjoint dataset which allows patient-agnostic feature extraction. We call this the ``offline phase'' as it only needs to be performed once.
The second part is an optional patient-specific fine-tuning and inference, which we refer to as the ``online phase''. Here, additional patient-specific details and anomalies can be learned.
During both phases, no \ac{hr} ground truth data is available. 
We optimize our model by using a loss function based on sparse continuous coordinates. Therefore, two orthogonal \ac{lr} images of the same object are viewed as sparse representations of the same underlying instance. 
Instead of fixed voxel coordinates, we compute the loss for suitable locations in the continuous coordinate space where the coordinates are normalized to the range [-1, 1]. This enables the seamless \ac{sr} of data with varying resolutions, as demonstrated in our ablation studies.
The applied loss computes the mean squared error ($\mathcal{L}\textsubscript{MSE}$) of the voxel intensity of reference \ac{lr} images and the voxel intensity of the predicted image $\hat{I}$ at matching coordinates $M_\mathrm{ax}$ and $M_\mathrm{cor}$,  where
\begin{equation}
    \mathcal{M_\mathrm{ax}}=\mathbf{C}_{\mathrm{HR}} \cap \mathbf{C}_{\mathrm{ax}} \text{  , and  } \mathcal{M_\mathrm{cor}}=\mathbf{C}_{\mathrm{HR}} \cap \mathbf{C}_{\mathrm{cor}} .
\end{equation}
Here, $\cap$ represents the intersection of available normalized coordinates.

Coordinates corresponding to regions without measurements in the \ac{lr} image are excluded from the loss calculation.

\section{Experiments}

\subsection{Datasets}
To assess performance, we simulated \ac{lr} images from \ac{hr} images and compared the \ac{sr} results to the corresponding \ac{hr}. We emphasize that the \ac{hr} data was used only as a ground truth for evaluation and was not used during model training.
We used two publicly available datasets for our evaluation: The Brain Tumor Segmentation (BraTS) dataset~\cite{Bakas.2017,Menze.2015}, consisting of brain \ac{mr} images with an isotropic resolution of 1$\times$1$\times$1\,mm\textsuperscript{3} and the HCP dataset~\cite{vanEssen.2013}, which includes 1200 \ac{mr} images with an isotropic resolution of 0.7$\times$0.7$\times$0.7\,mm\textsuperscript{3}.
For each dataset and each \ac{mr} sequence, we randomly selected 170 subjects (100 for training, 30 for validation, and 40 for final testing).

\subsection{Preprocessing}
For fair comparison, all images were registered using the NMI152 template with an isotropic resolution of 1\,mm. Background areas were removed, and images were normalized to $[0,1]$. 
To generate the anisotropic images in the \textit{training dataset}, we randomly downsampled each image in both axial and coronal direction using a random downsampling scale in the range $[2, 4]$ once at the beginning of the training.

We applied a commonly used downsampling method where \ac{hr} images are downsampled by cropping in the frequency domain~\cite{Li.2023,Masutani.2020}. 

\textit{For evaluation}, we generated two anisotropically sampled \ac{lr} \ac{mr} images in the axial and coronal dimension each with an in-slice resolution two and four times the between-slice resolution. 
As the \ac{lr} images were simulated from one \ac{hr} scan, they can be assumed to be rigidly registered. The axial images were resampled to a resolution of 1$\times$1$\times$f\,mm\textsuperscript{3}, while the coronal images were resampled to a resolution of 1$\times$f$\times$1\,mm\textsuperscript{3} with $\text{f}\in [2,4]$. 
All \ac{lr} image coordinates were normalized to a reference frame in the range $[-1,1]$.

\subsection{Implementation Details}

During the offline phase, we trained our network for 35 epochs with a batch size $B$ of 10 (lr=$0.0001$, Adam optimizer with $\beta_{1}=0.9$ and $\beta_{2}=0.999$), with hyperparameters tuned on the validation set. Training and evaluation were executed on an NVIDIA A40 with Python 3.9 and PyTorch 1.13.1. 
In each batch, $B$ randomly cropped \ac{lr} patches $p$ of one image were selected where $B$ equals the training batch size. The axial and coronal image patches were of size $10\times( e_i\times e_i\times 1)$ and $10\times( e_i\times 1\times e_i)$, respectively. To enable efficient batch processing for different upsampling scales, 8000 random samples are selected from each patch. Each sample corresponds to one \ac{hr} voxel coordinate. 
For the online phase, we used the trained model from the offline phase. For the optional patient-specific training, the model undergoes 10 additional epochs on \ac{lr} patches from a single patient before the final inference.

During the online phase, two different settings were evaluated. First, patient-specific online training was performed to update the model before subsequent inference. Second, inference without additional training was conducted which is referred to ``ours w/o FT''. We quantitatively evaluated models using Peak Signal-to-Noise Ratio (PSNR) and Structural Similarity Index Measure (SSIM)~\cite{Wang.2004}, which assess pixel accuracy and perceptual quality. To ensure a fair comparison, we computed the metrics only within the brain region and exclude the background, as BISR is only trained on the brain area.

\subsection{Reference Approaches}
As we trained our models without \ac{hr} ground truth data, we performed evaluations by comparing them with other unsupervised/self-supervised methods. We used cubic spline interpolation, a \ac{sisr} method SMORE~\cite{Zhao.2021} and an \ac{inr} \ac{misr} approach referred to as BISR~\cite{McGinnis.2023}. Originally,~\cite{McGinnis.2023} used two sequences as input and optimized the \ac{sr} jointly. We adapted the approach to directly predict only one \ac{sr} image instead of two images with different \ac{mr} sequences. Cubic spline interpolation and the \ac{sisr} method SMORE~\cite{Zhao.2021} are performed individually on the axial and coronal \ac{lr} images of each patient. The coronal and axial metric values are then averaged. 

\section{Quantitative and Qualitative Results}

\begin{table}[t]
{
\fontsize{8}{9}
\caption{Quantitative results for all \ac{mr} sequences and \ac{sr} methods on the BraTS and the HCP test set (trained and evaluated on the same \ac{mr} sequence). Best results are bold, second best underlined. ``Ours w/o FT'' refers to results without online training.}
\begin{center}
\begin{tabular}{
p{0.03\textwidth}>{\centering}
p{0.015\textwidth}>{\centering}
p{0.16\textwidth}>{\centering}
p{0.19\textwidth}>{\centering}
p{0.17\textwidth}|>{\centering}
p{0.19\textwidth}>{\centering\arraybackslash}
p{0.17\textwidth}} 
\hline
&\multicolumn{2}{c}{Resampling Scale} & \multicolumn{2}{c|}{$\times$2} & \multicolumn{2}{c}{$\times$4}\\
\hline

& & & PSNR $\uparrow$ & SSIM $\uparrow$ & PSNR $\uparrow$ & SSIM $\uparrow$ \\ 
\hline
\parbox[t]{2mm}{\multirow{15}{*}{\rotatebox[origin=c]{90}{BraTS}}
}
&\parbox[t]{2mm}{\multirow{5}{*}{\rotatebox[origin=c]{90}{T1 CE}}
}
& cubic spline & 33.794 $\pm$ 2.408 & 0.981 $\pm$ 0.006 
& 29.93 $\pm$ 2.285 & 0.945 $\pm$ 0.01\\ 
& & SMORE & 37.939 $\pm$ 1.483 & 0.981 $\pm$ 0.003 
& 30.678 $\pm$ 1.348 & 0.915 $\pm$ 0.009\\ 

& & BISR & \underline{42.727 $\pm$ 2.576} & \underline{0.993 $\pm$ 0.003}
& 35.008 $\pm$ 2.183 & 0.963 $\pm$ 0.007\\ 

& & Ours & \textbf{43.166 $\pm$ 2.105} & \textbf{0.994 $\pm$ 0.001}
& \textbf{35.598 $\pm$ 1.813} & \textbf{0.966 $\pm$ 0.005}\\ 

& & Ours w/o FT & 41.358 $\pm$ 2.701 & \underline{0.993 $\pm$ 0.002}
& \underline{35.531 $\pm$ 1.4655} & \underline{0.964 $\pm$ 0.004}\\ 
\cline{2-7}
& \parbox[t]{2mm}{\multirow{5}{*}{\rotatebox[origin=c]{90}{T1}}}
& cubic spline & 35.774 $\pm$ 3.745 & 0.991 $\pm$ 0.004 
& 29.624 $\pm$ 3.629 & 0.962 $\pm$ 0.009\\ 

& & SMORE & 34.333 $\pm$ 2.64 & 0.982 $\pm$ 0.003
& 26.627 $\pm$ 1.817 & 0.906 $\pm$ 0.008\\

& & BISR & \textbf{39.003 $\pm$ 3.984} & 0.993 $\pm$ 0.006
& \underline{31.078 $\pm$ 4.378} & \underline{0.967 $\pm$ 0.017} \\

& & Ours & \underline{38.832 $\pm$ 4.121} & \textbf{0.995 $\pm$ 0.002}
& \textbf{31.169 $\pm$ 3.33} & \textbf{0.969 $\pm$ 0.006}\\
& &  Ours w/o FT & 37.780 $\pm$ 4.061 & \underline{0.993 $\pm$ 0.003}
& 30.473 $\pm$ 3.632 & 0.965 $\pm$ 0.009\\ 
\cline{2-7}
& \parbox[t]{2mm}{\multirow{5}{*}{\rotatebox[origin=c]{90}{T2}}}
& cubic spline & 33.890 $\pm$ 1.876 & 0.986 $\pm$ 0.004 
& 30.643 $\pm$ 2.168 & 0.959 $\pm$ 0.014\\ 

&& SMORE & 35.645 $\pm$ 1.388 & 0.984 $\pm$ 0.004
& 27.799 $\pm$ 1.217 & 0.91 $\pm$ 0.015\\

& & BISR & \textbf{42.460 $\pm$ 1.853} & \textbf{0.997 $\pm$ 0.001}
& \textbf{34.74 $\pm$ 1.661} & \textbf{0.976 $\pm$ 0.007}\\

& &Ours & \underline{42.218 $\pm$ 1.750} & \textbf{0.997 $\pm$ 0.001}
& \underline{33.427 $\pm$ 1.640} & \underline{0.973 $\pm$ 0.009}\\

& &  Ours w/o FT & 40.531 $\pm$ 2.190 & 0.995 $\pm$ 0.002
& 33.073 $\pm$ 1.412 & 0.969 $\pm$ 0.008\\ 
\hline

\parbox[t]{2mm}{\multirow{10}{*}{\rotatebox[origin=c]{90}{HCP}}
}
&
\parbox[t]{2mm}{\multirow{5}{*}{\rotatebox[origin=c]{90}{T1}}}
 & cubic spline & 29.820 $\pm$ 3.785 & 0.982 $\pm$ 0.006 
& 22.341 $\pm$ 3.947 & 0.930 $\pm$ 0.023\\ 
& & SMORE & 30.603 $\pm$ 4.099 & 0.973 $\pm$ 0.007
& 22.533 $\pm$ 3.177 & 0.881 $\pm$ 0.019\\

& & BISR & \underline{31.249 $\pm$ 4.574} & \underline{0.985 $\pm$ 0.007}
& \underline{22.758 $\pm$ 4.636} & 0.931 $\pm$ 0.034 \\

& & Ours & \textbf{34.140 $\pm$ 5.410} & \textbf{0.990 $\pm$ 0.004}
& \textbf{23.301 $\pm$ 5.274} & \textbf{0.938 $\pm$ 0.025}\\

& &  Ours w/o FT & 29.950 $\pm$ 4.655 & 0.984 $\pm$ 0.007
& 22.523 $\pm$ 4.860 & \underline{0.933 $\pm$ 0.026}\\ 

\cline{2-7}
& \parbox[t]{2mm}{\multirow{5}{*}{\rotatebox[origin=c]{90}{T2}}}
 & cubic spline & 30.561 $\pm$ 1.275 & 0.973 $\pm$ 0.004 
& 27.336 $\pm$ 0.864 & 0.930 $\pm$ 0.008\\ 
& & SMORE & 31.444 $\pm$ 0.814 & 0.967 $\pm$ 0.004
& 25.817 $\pm$ 0.597 & 0.877 $\pm$ 0.010\\

& & BISR & \underline{35.321 $\pm$ 0.858} & \underline{0.985 $\pm$ 0.002}
& \textbf{29.862 $\pm$ 0.531} & \textbf{0.947 $\pm$ 0.004}\\

& & Ours & \textbf{35.482 $\pm$ 0.877} & \textbf{0.986 $\pm$ 0.001}
& 29.372 $\pm$ 1.231 & 0.941 $\pm$ 0.006\\

& & Ours w/o FT & 34.841 $\pm$ 1.155 & \underline{0.985 $\pm$ 0.002}
& \underline{29.681 $\pm$ 1.055} & \underline{0.942 $\pm$ 0.006}\\ 

\end{tabular}
\end{center}
\label{table_results1_brats}
}
\end{table}

\begin{table}[t]
\caption{Duration of the offline and online phase for each method in minutes. Best, \ie, shortest times are highlighted in bold.}
\begin{center}
\begin{tabular}{
p{0.02\textwidth}>{\centering}
p{0.02\textwidth}>{\centering}
p{0.18\textwidth}>{\centering}
p{0.16\textwidth}>{\centering}
p{0.18\textwidth}|>{\centering}
p{0.16\textwidth}>{\centering\arraybackslash}
p{0.18\textwidth}} 
\hline
& & Resampling Scale & \multicolumn{2}{c|}{$\times$2} & \multicolumn{2}{c}{$\times$4}\\
\hline

& & & offline & online & offline & online \\
\hline
\parbox[t]{2mm}{\multirow{4}{*}{\rotatebox[origin=c]{90}{BraTS}}}
&\parbox[t]{2mm}{\multirow{4}{*}{\rotatebox[origin=c]{90}{T1}}
}
& SMORE & 0 &  9.09 $\pm$ 0.29
& 0 & 9.16 $\pm$ 0.32\\ 
& & BISR & 0 & 19.25 $\pm$ 1.32 
& 0 &  10.59 $\pm$ 0.67\\ 
& & Ours & 130.99 $\pm$ 5.15 & 3.32 $\pm$ 0.31
& 130.99 $\pm$ 5.15 & 1.55 $\pm$ 0.15 \\ 
& & Ours w/o FT & 130.99 $\pm$ 5.15 & \textbf{1.94 $\pm$ 0.22}
& 130.99 $\pm$ 5.15 & \textbf{0.73 $\pm$ 0.08} \\ 
\hline
\end{tabular}
\end{center}
\label{table_timing}
\end{table}

\begin{table}[t]
{
\fontsize{8}{9}
\caption{Quantitative results for different training settings with patient-specific online training. ``Train on''  refers to the used training data and ``test on'' refers to the specific test set. ``Br'' is used as abbreviation for BraTS.}
\begin{center}
\begin{tabular}{
p{0.12\textwidth}>{\centering}
p{0.12\textwidth}>{\centering}
p{0.18\textwidth}>{\centering}
p{0.17\textwidth}|>{\centering}
p{0.18\textwidth}>{\centering\arraybackslash}
p{0.17\textwidth}} 
\hline
\multicolumn{2}{c}{Resampling Scale} & \multicolumn{2}{c|}{$\times$2} & \multicolumn{2}{c}{$\times$4}\\
\hline

Train on & Test on & PSNR $\uparrow$ & SSIM $\uparrow$ & PSNR $\uparrow$ & SSIM $\uparrow$ \\ 
\hline
Br T1 CE & Br T1 CE & 43.166 $\pm$ 2.105 & 0.994 $\pm$ 0.001
& 35.598 $\pm$ 1.813 & 0.966 $\pm$ 0.005\\ 

Br T1 & Br T1 CE & 42.680 $\pm$ 2.435 & 0.994 $\pm$ 0.001
& 35.275 $\pm$ 1.818 & 0.963 $\pm$ 0.005\\ 

Br T2 & Br T1 CE & 42.762 $\pm$ 2.285 & 0.994 $\pm$ 0.001 
& 35.476 $\pm$ 1.66 & 0.963 $\pm$ 0.005\\ 

\cline{1-6}
Br T1 & Br T1 & 38.832 $\pm$ 4.121 & 0.995 $\pm$ 0.002
& 31.169 $\pm$ 3.33 & 0.969 $\pm$ 0.006\\

Br T1 CE & Br T1 & 40.178 $\pm$ 4.388 & 0.995 $\pm$ 0.003 
& 31.457 $\pm$ 3.382 & 0.970 $\pm$ 0.006\\ 

Br T2& Br T1 &39.981 $\pm$ 4.259 & 0.995 $\pm$ 0.002
& 31.474 $\pm$ 3.245 & 0.969 $\pm$ 0.007\\

HCP T1 & Br T1 & 39.410 $\pm$ 4.200 & 0.995 $\pm$ 0.002
& 31.450 $\pm$ 3.674 & 0.970 $\pm$ 0.006\\ 

\cline{1-6}
Br T2 & Br T2 & 42.218 $\pm$ 1.750 & 0.997 $\pm$ 0.001
& 33.427 $\pm$ 1.640 & 0.973 $\pm$ 0.009\\

Br T1 CE & Br T2 &  42.360 $\pm$ 1.962 & 0.997 $\pm$ 0.001 
& 34.290 $\pm$ 1.359 & 0.976 $\pm$ 0.007\\ 

Br T1 & Br T2 & 42.226 $\pm$ 1.958 & 0.997 $\pm$ 0.002
& 33.610 $\pm$ 1.467 & 0.973 $\pm$ 0.008\\

HCP T2 & Br T2 & 42.456 $\pm$ 2.311 & 0.997 $\pm$ 0.002
& 33.548 $\pm$ 1.819 & 0.975 $\pm$ 0.009\\ 

\hline
\end{tabular}
\end{center}
\label{table_results1_mixed}
}
\end{table}

We evaluated on 40 brain \ac{mr} images each for T1, T1 \ac{ce}, and T2 sequences for BraTS and T1 and T2 for HCP which are trained either on images from the same sequence or from any of the other sequences (for $2\times$ and $4\times$ upsampling).  Table~\ref{table_results1_brats} summarizes the quantitative results for all \ac{sr} approaches including mean and standard deviation. We achieve higher or comparable PSNR and SSIM values on all \ac{mr} sequences and upsampling scales. Using the additional patient-specific online training helps to improve the performance in all cases. Still, without additional training (ours w/o FT), the performance is comparable to the best reference approach at approx.\ 10-15$\times$ faster inference times. Table~\ref{table_results1_mixed} shows results obtained from testing on a dataset distinct from the one used for training. Even when the network is trained on a completely different dataset, the \ac{sr} performance is comparable or even better compared to training on the same dataset.

The time required to generate an \ac{sr} reconstruction for each method can be divided into offline and online phases. The results are shown in Table~\ref{table_timing}. Our model is pre-trained once offline which takes approximately 130 minutes. We average the time for the offline phase for three runs.
During the online phase, our approach is approximately 10-15$\times$ faster when using the offline model w/o patient-specific fine-tuning and still 6$\times$ faster when fine-tuning for a specific patient, compared to the reference approach BISR.

\begin{figure}[t]
\includegraphics[width=\textwidth]{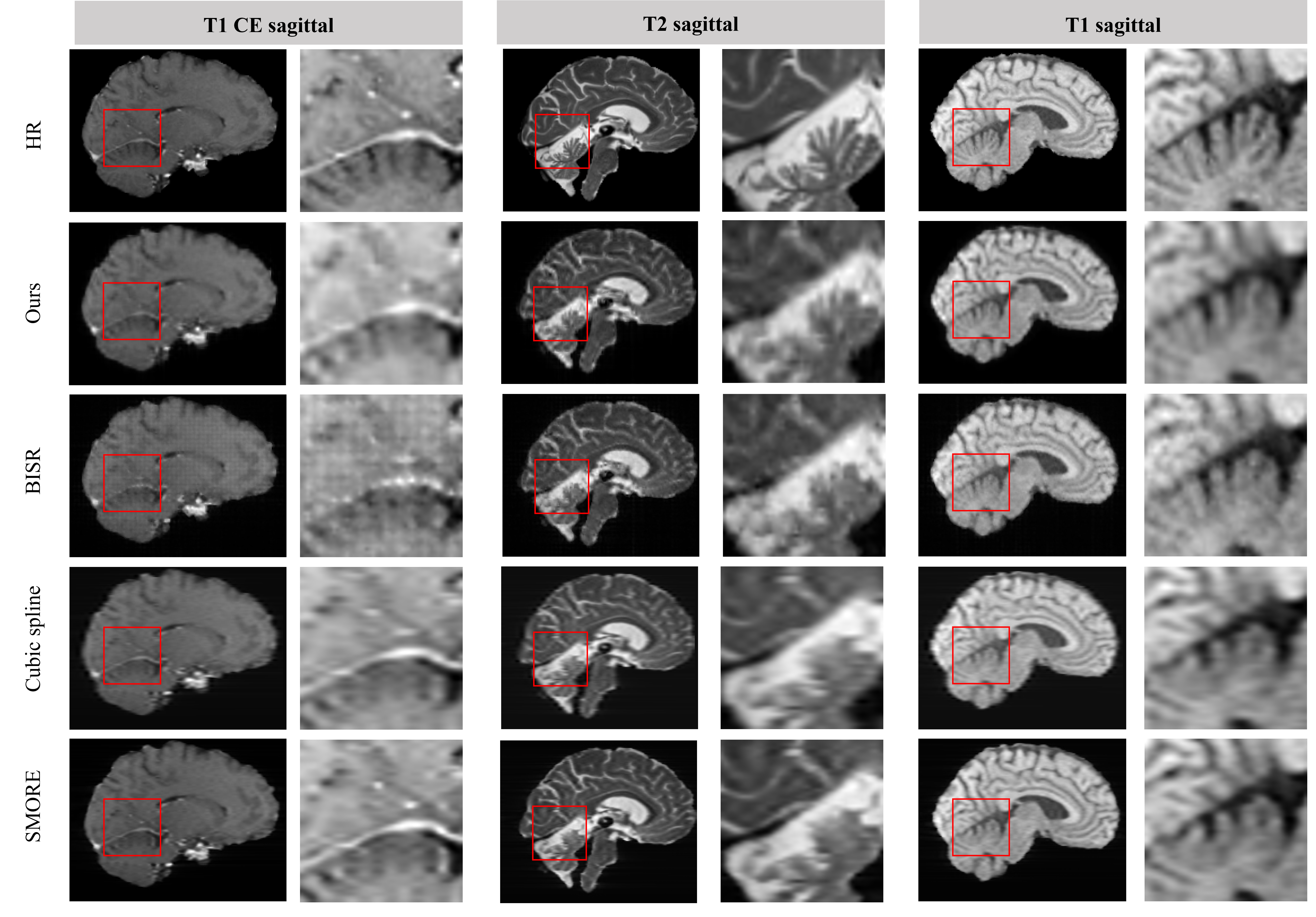}
\caption{Reference \ac{hr} image and qualitative \ac{sr} results for all \ac{mr} sequences in the sagittal plane of the BraTS test set where no in-plane \ac{hr} images are available.} \label{quant_fig}
\end{figure}

In~\cref{quant_fig}, qualitative examples of the BraTS dataset for each \ac{mr} sequence and \ac{sr} method can be seen in comparison to the actual \ac{hr} image. The full image and a zoomed-in version can be seen for each sample. Overall, the \ac{sr} results obtained using SMORE and cubic spline interpolation appear blurred or noisy. In some cases, BISR introduces noticeable blocking artifacts. In contrast, our method preserves accurate anatomical structures while minimizing noise.

\section{Discussion and Conclusion}
In this work, we presented a novel approach for \ac{mr} image \ac{sr}. 
In both quantitative and qualitative evaluations, our method produces \ac{hr} images that match or surpass the quality of current reference approaches. A key strength lies in its substantially reduced patient-specific training time during the online phase, which is critical for translating such techniques into clinical practice. Notably, after generating only 14 \ac{sr} images with a resampling factor of 4 using BISR, the time invested in our offline phase is already offset. This efficiency gain enables a ten-fold increase in throughput for the generation of \ac{sr} data. This could help to further drive acceptance of deep learning-supported diagnostic imaging in clinical workflows. Additionally, the sparse coordinate loss enables the integration of \ac{lr} images with varying resolution scales, further enhancing the versatility and applicability of our approach.

Finally, the presented method demonstrates promising capabilities by consistently achieving high-quality results, regardless of whether training is performed on data from the same cohort or on a disjoint dataset. 
Further experiments are needed to validate our findings and demonstrate the benefits of using the reconstructed \ac{sr} images in diagnostic assessments.

\begin{credits}
\subsubsection{\ackname} We gratefully acknowledge support by d.hip campus - Bavarian aim (Ma.S. and K.B.) and HPC resources provided by the Erlangen National High Performance Computing Center (NHR@FAU) of the Friedrich-Alexander-Universität Erlangen-Nürnberg (FAU).

\subsubsection{\discintname}
The authors have no competing interests to declare that are relevant to the content of this article.
\end{credits}

\begin{acronym}
\acro{wsi}[WSI]{whole slide image}
\acro{mr}[MR]{Magnetic Resonance}
\acro{sr}[SR]{super-resolution}
\acro{snr}[SNR]{signal-to-noise ratio}
\acro{lr}[LR]{low-resolution}
\acro{hr}[HR]{high-resolution}
\acro{misr}[MISR]{multi-image super-resolution}
\acro{sisr}[SISR]{single-image super-resolution}
\acro{inr}[INR]{implicit neural representation}
\acro{mlp}[MLP]{multi-layer perceptron}
\acro{ce}[CE]{contrast-enhanced}
\acro{ct}[CT]{Computed Tomography}
\acro{inr}[INR]{implicit neural representation}
\acro{sota}[SOTA]{state-of-the-art}
\acro{fft}[FFT]{Fast Fourier transform}
\acro{us}[US]{unsupervised/ self-supervised}
\acro{sv}[SV]{supervised}
\acro{rdn}[RDN]{residual dense network}

\end{acronym}

\bibliographystyle{splncs04}
\bibliography{multi_view_sr.bib}

\end{document}